\documentclass[conference]{IEEEtran}
\IEEEoverridecommandlockouts
\usepackage{cite}
\usepackage{amsmath,amssymb,amsfonts}
\usepackage{algorithmic}
\usepackage{graphicx}
\usepackage{textcomp}
\usepackage{color}
\usepackage{multirow}
\usepackage{comment}
\usepackage{xcolor}
\usepackage{hyperref}
\hypersetup{
    colorlinks=true,
    linkcolor=blue,
    filecolor=magenta,      
    urlcolor=cyan,
    pdftitle={Overleaf Example},
    pdfpagemode=FullScreen,
    }
\def\BibTeX{{\rm B\kern-.05em{\sc i\kern-.025em b}\kern-.08em
    T\kern-.1667em\lower.7ex\hbox{E}\kern-.125emX}}
\begin{document}

\title{Referenceless User Controllable Semantic Image Synthesis}

\author{
\IEEEauthorblockN{Jonghyun Kim}
\IEEEauthorblockA{\textit{AI Lab, CTO Division} \\
\textit{LG Electronics}\\
Seoul, Republic of Korea \\
jonghyun0.kim@lge.com, jhkim.ben@gmail.com}
\and
\IEEEauthorblockN{Gen Li}
\IEEEauthorblockA{\textit{School of Informatics} \\
\textit{University of Edinburgh}\\
Edinburgh, UK \\
li.gen@ed.ac.uk}
\and
\IEEEauthorblockN{Joongkyu Kim* \thanks{*Corresponding author}}
\IEEEauthorblockA{\textit{Dept. of Electrical and Computer Engineering} \\
\textit{Sungkyunkwan University}\\
Suwon, Republic of Korea \\
jkkim@skku.edu}
}

\maketitle

\begin{abstract}
Despite recent progress in semantic image synthesis, complete control over image style remains a challenging problem. Existing methods require reference images to feed style information into semantic layouts, which indicates that the style is constrained by the given image. In this paper, we propose a model named RUCGAN for user controllable semantic image synthesis, which utilizes a singular color to represent the style of a specific semantic region. The proposed network achieves reference-free semantic image synthesis by injecting color as user-desired styles into each semantic layout, and is able to synthesize semantic images with unusual colors. Extensive experimental results on various challenging datasets show that the proposed method outperforms existing methods, and we further provide an interactive UI to demonstrate the advantage of our approach for style controllability. The codes and UI are available at: \url{https://github.com/BenjaminJonghyun/RUCGAN}
\end{abstract}


\section{Introduction}
Conditional generative adversarial networks (cGAN) \cite{mirza2014conditional} have become popular for generating images with conditional settings. Many efforts have been made to generate realistic images conditioning on latent codes \cite{brock2018large,zhang2019self}, edge maps \cite{alharbi2019latent}, other images \cite{isola2017image}, pose key points \cite{tang2020bipartite}, and texts \cite{zhang2017stackgan}. In particular, we focus on conditional image synthesis with semantic layouts \cite{wang2018high,park2019semantic,zhu2020sean,schonfeld2020you,le2021semantic,tan2021efficient}, where images are generated from segmentation maps. According to the image style control, existing work can be categorized into two parts: random style generation and user-specified style generation. The former ones \cite{isola2017image,schonfeld2020you} generate synthesized images using semantic layouts with random noise, therefore image styles cannot be freely manipulated. To make the image style controllable, the latter ones \cite{wang2018high,park2019semantic,le2021semantic,tan2021efficient} utilize features or distributions extracted from reference images. Although these methods can render style-duplicated images, they are constrained in global manipulation. To control the style of each segmentation label individually, SEAN \cite{zhu2020sean} and SuperStyleNet \cite{kim2021superstylenet} encode style vectors per semantic region, and utilize them to learn spatially varying normalization parameters for style reconstruction. However, there are some drawbacks in these methods that lead to undesirable contents in the process of style extraction.
First, dozens of reference images are required to extract desired styles. Assuming we extract styles for each semantic region from different images, then reference images should be equal to the number of layout classes, and it is time-consuming to find preferred images among the reference ones. Second, there are limitations in the style extraction process. To encode styles from a reference image, SEAN multiplies feature maps extracted by the style encoder with its corresponding semantic layout. In this case, synthesized styles are restricted to images in a given dataset, which indicates that the image manipulation is unavailable when desired styles do not exist.

To tackle these shortcomings, a straightforward strategy is to extract user-specified styles from their thoughts. However, it is hard to convert them as input data for neural networks. Therefore, we focus on colors instead to simply represent styles of objects. In this paper, we present a novel generative adversarial network, named RUCGAN, that learns image styles from a color bank to achieve referenceless user controllable semantic image synthesis. Specifically, RUCGAN allows user-preferred styles without the help of reference images, and is capable of generating photo-realistic images with unusual styles.

We provide extensive experiments to prove the effectiveness of the proposed method on three challenging datasets: Landscapes High-Quality (HQ) \cite{ALIS}, CelebAMask-HQ \cite{lee2020maskgan}, and Cityscapes \cite{cordts2016cityscapes}, and evaluations are made with both segmentation and generation metrics. Compared with existing methods, our contributions can be summarized as follows:

\begin{itemize}
    \item We propose a novel architecture, named RUCGAN, to map colors into semantic layouts as style information, which can achieve data-unconstrained image synthesis in local and global regions without reference images.
    \item We introduce training strategies to help RUCGAN synthesize semantic images with data-unconstrained styles.
    \item We further provide an interactive UI to easily customize desired styles for each semantic layout using a color bank and generate realistic synthesized images.
\end{itemize}

\begin{figure*}[t]
    \centering
    \includegraphics[width=1.0\textwidth]{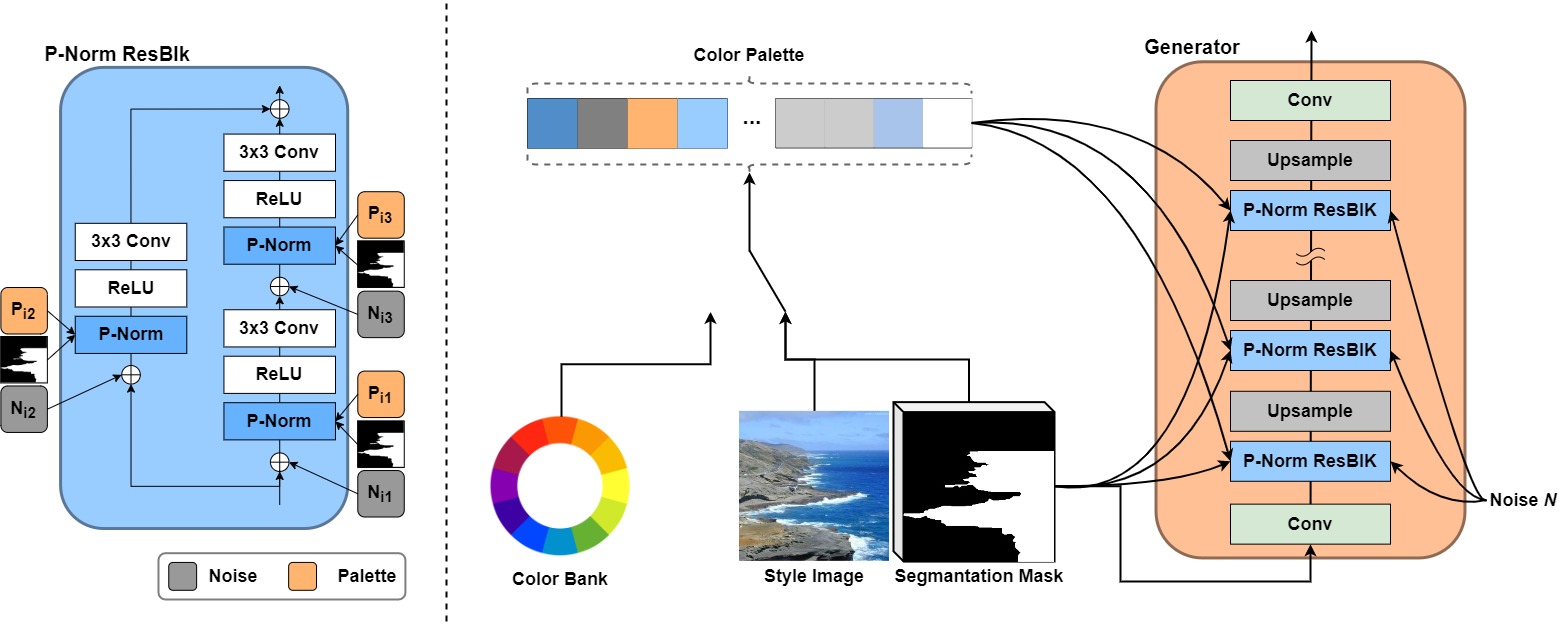}
    \caption{The overall framework of the proposed RUCGAN. {\textit{(left)}} Structure of P-Norm ResBlk. {\textit{(right)}} Generator of RUCGAN, which consists of a series of the P-Norm ResBlk with upsampling layers. The color palette can be obtained from the style image or the color bank.}
    \label{architecture}
\end{figure*}

\section{Related work}

\subsection{Semantic image synthesis}
Isola {\it et al.} \cite{isola2017image} first proposed image-to-image translation with conditional adversarial networks \cite{mirza2014conditional} to translate input data to target images. As a specific task of it, semantic image synthesis
\cite{wang2018high,liu2019learning,schonfeld2020you,wang2021image} generates realistic images from semantic layouts. To be specific, wang {\it et al.} \cite{wang2018high} utilized multi-scale encoder-decoder structures to generate high-resolution synthesized images. Liu {\it et al.} \cite{liu2019learning} predicted convolutional kernel weights based on the semantic layout to generate synthesized images by adopting them into random noise. Similarly, SC-GAN \cite{wang2021image} encoded semantic information as semantic vectors, and spatially normalized random noise input using them. Aforementioned methods improved a generator to synthesize high-quality images. On the other hand, OASIS \cite{schonfeld2020you} refined a discriminator which segments pixels of target and synthesized images into real and fake classes. Although these methods change image styles using random noise, they are globally and locally uncontrollable.

\subsection{Style controllability}
To solve the above issue, existing methods \cite{park2019semantic,zhu2020sean,le2021semantic,tan2021efficient,tan2021diverse,kim2021superstylenet} extracted styles from reference images, and injected them into semantic layouts. To be specific, SPADE \cite{park2019semantic} proposed spatially-adaptive normalization to recover information contained in the input semantic masks. Moreover, CLADE \cite{tan2021efficient} reduced learnable normalization parameters of SPADE by using a modulation parameter bank. Although these methods generate high-quality images from semantic layouts, output image styles can only be manipulated  globally. To achieve local image editing, SEAN \cite{zhu2020sean} and INADE \cite{tan2021diverse} extracted style features using encoder-decoder networks, and injected them into each semantic layout by adopting semantic region-adaptive or instance-adaptive normalization. Different from these methods, SuperStyleNet \cite{kim2021superstylenet} utilized superpixel algorithms \cite{achanta2012slic,irving2016maskslic} to extract style vectors from a reference image. However, such image editing requires bags of time to pick out desired style images for each semantic region, and reconstructed styles are constrained to the images of a given dataset. To overcome these shortcomings, we not only reduce image selection steps by providing color palettes but also generate realistic images with various colors.

\begin{figure*}[t]
    \centering
    \includegraphics[width=1.0\textwidth]{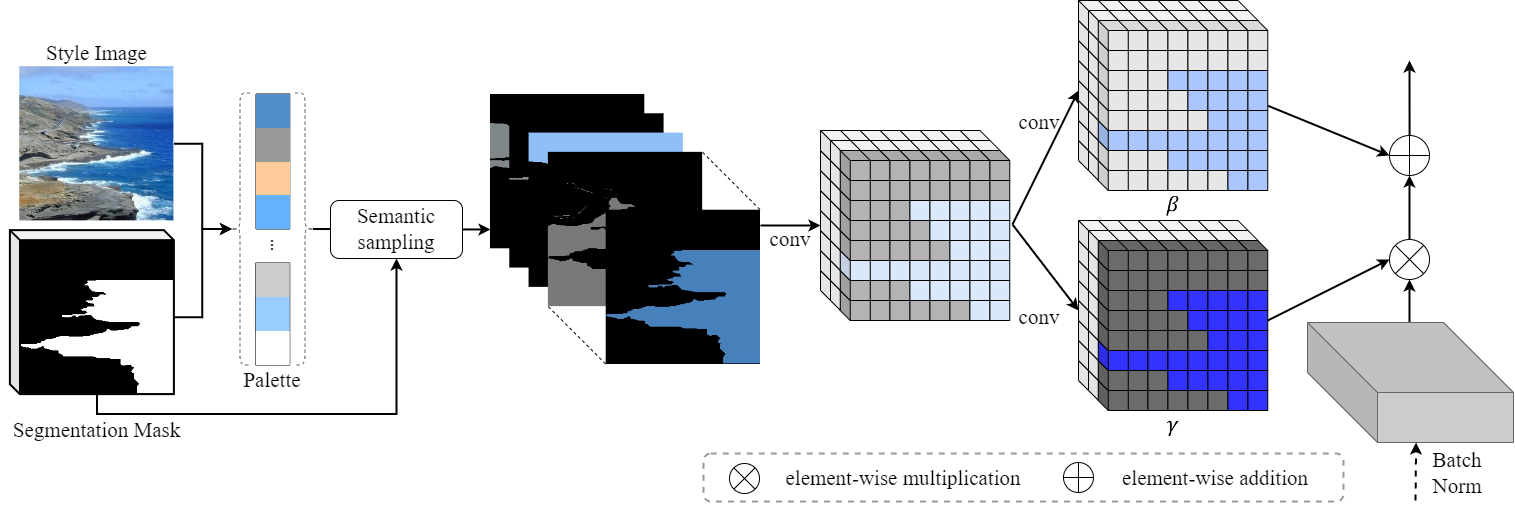}
    \caption{The illustration of palette normalization (P-Norm). First, representative colors in each semantic label are extracted from the style image. After that, these colors are semantically sampled and injected into corresponding semantic labels. Finally, two modulation parameters $\gamma, \beta$ are yielded from shallow convolution layers to normalize input features.}
    \label{pnorm}
\end{figure*}

\begin{figure*}[t]
    \centering
    \includegraphics[width=1.0\textwidth]{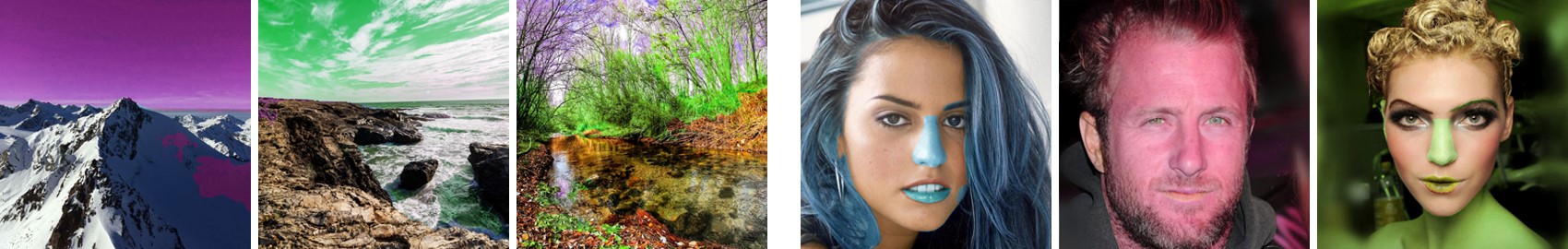}
    \caption{Examples of semantic color jitter on Landscapes High-Quality and CelebAMask-HQ. The hue component of style images is stochastically transformed.}
    \label{colorjitter}
\end{figure*}

\section{Proposed method}
Our main objective is to represent user-desired styles as visible data. To solve this issue, we focus on colors of each semantic region since most stuff in the semantic regions has their specific colors, i.e., blue sky and green vegetation. Therefore, we propose RUCGAN to learn styles of objects from colors using semantic modulation parameters as shown in \ref{architecture}. Specifically, palette normalization (P-Norm) allows that RUCGAN generates realistic styles of each semantic region employing colors extracted from reference images and selected by users from the color bank. Furthermore, semantic color mix (SCM) facilitates RUCGAN to synthesize unusual cases.

In this section, we first introduce palette normalization (P-Norm) and learning strategies. Thereafter, we describe the overall architecture of RUCGAN.

\subsection{Palette normalization}
To convert user thought to visible data, we adopt a singular color to represent the style of a specific semantic region, and the representative color is obtained by averaging pixel values of each segmentation mask. By doing this, we extract a $s \times 1$ palette vector $P$ from a style image with its semantic segmentation mask, where $s$ is the number of semantic labels. The palette vector is semantically sampled into semantic layout $M$ to inject it into each semantic region as shown in Figure \ref{pnorm}. Then, the outputs of semantic sampling are concatenated along the channel dimension to obtain a semantic style map, which allows each channel to contain semantically independent style information. Afterward, two separate convolutional layers are used to learn two modulation parameters $\gamma$ and $\beta$ from the semantic style map, respectively. Finally, the input features of this layer are normalized across channels and modulated with two learned modulation parameters.
This method allows users to simply control styles in each segmentation mask without resorting to reference images.

Let $h$ denote the activations of a RUCGAN generator for a batch of $N$ samples. Let $H, W$, and $C$ be the height, width, and the number of channels in the activation map. The activation value at site ($n \in N, c \in C, y \in H, x \in W$) is
\begin{equation}
    \gamma_{c,y,x}(P,M)\frac{h_{n,c,y,x}-\mu_{c}}{\sigma_c}+\beta_{c,y,x}(P,M),
\end{equation}
where $h_{n,c,y,x}$ is the activation before normalization, and $\mu_{c}$ and $\sigma_{c}$ are its mean and standard deviation in channel c:
\begin{equation}
    \mu_{c}=\frac{1}{NHW}\sum_{n,y,x}h_{n,c,y,x}
\end{equation}
\begin{equation}
    \sigma_{c}=\sqrt{\frac{1}{NHW}\sum_{n,y,x}((h_{n,c,y,x})^2-(\mu_c)^2)}.
\end{equation}
$\gamma_{c,y,x}(P,M)$ and $\beta_{c,y,x}(P,M)$ are modulation parameters learned from the semantic style map.

\subsection{Semantic color mix}
The main obstacle to this task is mapping unusual colors to specific semantic regions, i.e., green sky and purple ocean. As colors of semantic regions are constrained by dataset, therefore it is difficult to synthesize realistic images when users select unusual cases. To solve this problem, we apply color jitter to style images during the training phase. In this procedure, we exclude brightness, contrast, and saturation options for color jitter, since user-selected colors are not intactly reflected. For example, light tones affect color jitter of all options simultaneously. Thereby we only adopt the hue transformation in the color jitter to accurately represent the selected colors in semantic regions. To be specific, the hue of the style images varies randomly with the maximum displacement of the jitter parameter\footnote{This parameter is set to $[-0.5, 0.5]$ in a $torchvision$ tool}.

Furthermore, we propose semantic color mix to prevent the color change of a semantic label from affecting other labels. In the training phase, we conduct color jitter on randomly given semantic layouts. Let $L$ be the number of labels given a semantic mask $M$, and $\frac{L}{2}$ labels are randomly selected from the semantic mask. Then, we generate a set of integers $\mathbb{M}$, which consists of ``1'' if $M == l \in \frac{L}{2}$ otherwise ``0''. After that, we conduct semantic color mix as follows:
\begin{equation}
    \hat{X}=X\otimes \mathbb{M} + X_{c} \otimes (1-\mathbb{M}),
    \label{jitter}
\end{equation}
where $X$ and $X_{c}$ are the original image and the corresponding color jittered image, respectively. $\otimes$ denotes pixel-wise multiplication. Examples of Equation \ref{jitter} are illustrated in Figure \ref{colorjitter}.

\begin{table*}[t]
\centering
\caption{Quantitative comparison with state-of-the-art methods. For mIoU, pixel-wise accuracy (acc), and style relevance (SR), higher is better. For FID and LPIPS, lower is better.}
\small
\begin{tabular}{l||ccc||ccccc||ccccc}
\hline
\multicolumn{1}{c||}{\multirow{2}{*}{Method}} & \multicolumn{3}{c||}{Landscape HQ}           & \multicolumn{5}{c||}{CelebAMask-HQ}                                                                                               & \multicolumn{5}{c}{Cityscapes}                                                                                                  \\ \cline{2-14} 
\multicolumn{1}{c||}{}                        & \multicolumn{1}{c|}{FID}            &  \multicolumn{1}{c|}{LPIPS} & SR           & \multicolumn{1}{c|}{mIoU}           & \multicolumn{1}{c|}{acc}            & \multicolumn{1}{c|}{FID}            & \multicolumn{1}{c|}{LPIPS} & SR          & \multicolumn{1}{c|}{mIoU}           & \multicolumn{1}{c|}{acc}            & \multicolumn{1}{c|}{FID}            & \multicolumn{1}{c|}{LPIPS} & SR          \\ \hline \hline
Pix2PixHD \cite{wang2018high}                                    & \multicolumn{1}{c|}{80.22}          & \multicolumn{1}{c|}{0.450} & 0.915           & \multicolumn{1}{c|}{73.15}          & \multicolumn{1}{c|}{95.22}          & \multicolumn{1}{c|}{27.45}          & \multicolumn{1}{c|}{0.359} & 0.849          & \multicolumn{1}{c|}{49.21}          & \multicolumn{1}{c|}{91.18}          & \multicolumn{1}{c|}{104.39}         & \multicolumn{1}{c|}{0.393} & 0.789          \\ \hline
SPADE \cite{park2019semantic}                                        & \multicolumn{1}{c|}{45.11}          & \multicolumn{1}{c|}{0.509} & 0.871          & \multicolumn{1}{c|}{\textbf{74.55}} & \multicolumn{1}{c|}{\textbf{95.72}} & \multicolumn{1}{c|}{33.94}          & \multicolumn{1}{c|}{0.373} & 0.827          & \multicolumn{1}{c|}{\textbf{55.02}} & \multicolumn{1}{c|}{92.93}          & \multicolumn{1}{c|}{\textbf{51.18}} & \multicolumn{1}{c|}{0.380} & 0.812          \\ \hline
SEAN \cite{zhu2020sean}                                         & \multicolumn{1}{c|}{18.79} & \multicolumn{1}{c|}{\textbf{0.285}} & 0.941 & \multicolumn{1}{c|}{72.63}          & \multicolumn{1}{c|}{95.28}          & \multicolumn{1}{c|}{22.41}          & \multicolumn{1}{c|}{0.274} & 0.870          & \multicolumn{1}{c|}{52.52}          & \multicolumn{1}{c|}{92.53}          & \multicolumn{1}{c|}{52.62}          & \multicolumn{1}{c|}{\textbf{0.331}} & 0.817 \\ \hline
SuperStyleNet \cite{kim2021superstylenet}                                & \multicolumn{1}{c|}{\textbf{18.61}}          & \multicolumn{1}{c|}{0.294} & \textbf{0.943}           & \multicolumn{1}{c|}{73.89}          & \multicolumn{1}{c|}{\textbf{95.72}} & \multicolumn{1}{c|}{25.49}          & \multicolumn{1}{c|}{\textbf{0.255}} & 0.869 & \multicolumn{1}{c|}{53.37}          & \multicolumn{1}{c|}{\textbf{93.01}} & \multicolumn{1}{c|}{60.45}          & \multicolumn{1}{c|}{0.349} & \textbf{0.825}          \\ \hline
\textbf{Ours}                                 & \multicolumn{1}{c|}{21.32}          & \multicolumn{1}{c|}{0.347} & \textbf{0.943}          & \multicolumn{1}{c|}{73.22}          & \multicolumn{1}{c|}{95.31}          & \multicolumn{1}{c|}{\textbf{21.80}} & \multicolumn{1}{c|}{0.282} & \textbf{0.872}          & \multicolumn{1}{c|}{50.74}          & \multicolumn{1}{c|}{91.34}          & \multicolumn{1}{c|}{76.39}          & \multicolumn{1}{c|}{0.438} & 0.817          \\ \hline
\end{tabular}
\label{quan_comp}
\end{table*}

\begin{figure*}[t]
    \centering
    \includegraphics[width=1.0\textwidth]{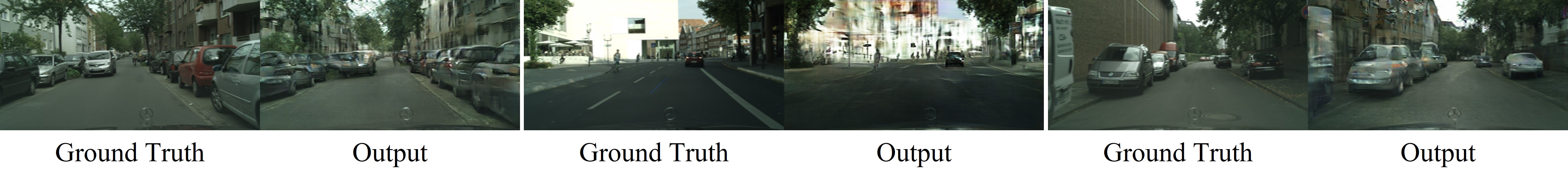}
    \vspace{-0.5cm}
    \caption{Failure cases on Cityscapes.}
    \label{fail}
\end{figure*}

\subsection{Network architecture}
Following existing methods \cite{kim2021superstylenet,tan2021diverse,tan2021efficient,zhu2020sean,park2019semantic}, we adopt similar network structures. As shown in Figure \ref{architecture}, the proposed network mainly consists of several P-Norm Residual blocks (ResBlk) with upsampling layers. Specifically, each P-Norm ResBlk contains three convolutional layers with three P-Norm layers. Before modulating activations in the P-Norm ResBlk, noise $N$ is added to the input of each P-Norm layer. For each P-Norm layer, a color palette and a segmentation mask are utilized. The color palette can be extracted from a style image with its corresponding semantic layout or selected from the color bank by users. In addition, we adopt a multi-scale discriminator \cite{wang2018high} to generate realistic images, and two types of input are fed into the discriminator: the concatenation of the segmentation mask with a synthesized image or a real image.

In order to train the generator of RUCGAN, we adopt three types of learning objectives: conditional adversarial loss, feature matching loss \cite{wang2018high}, and perceptual loss \cite{johnson2016perceptual}. For the multi-scale discriminator, the hinge loss \cite{miyato2018spectral,zhang2019self} is applied.

\section{Experiments}
\subsection{Experimental setting}
Following existing methods \cite{kim2021superstylenet,zhu2020sean}, learning rates are set to 0.0001 and 0.0004 for the generator and discriminator with a two time-scale update rule (TTUR) \cite{heusel2017gans}. To update loss functions, Adam optimizer \cite{kingma2014adam} is applied with $\beta_{1}=0.5$ and $\beta_{2}=0.999$. All experiments are conducted on 4 32G Tesla V100 GPUs.


\subsection{Dataset}
We conduct experiments on various datasets as follows:
\begin{itemize}
    \item \textit{Landscapes High-Quality} \cite{ALIS} contains 90,000 high-resolution landscape images, which are collected from Unsplash and Flickr. We adopt DeepLabV2 \cite{chen2017deeplab} pretrained on COCO-Stuff \cite{caesar2018coco} to generate input segmentation masks. The train and test set sizes are 85,000 and 5,000, respectively.
    \item \textit{CelebAMask-HQ} \cite{lee2020maskgan} consists of 30,000 face images with 19 segmentation labels. We divide this dataset into 28,000 and 2,000 images for train and test sets.
    \item \textit{Cityscapes} \cite{cordts2016cityscapes} contains 3,500 street scene images with 35 segmentation labels. This dataset is split into 3,000 and 500 for train and test sets.
\end{itemize}

\subsection{Evaluation metrics}
In order to quantitatively compare the proposed method with state-of-the-arts, we adopt metrics in semantic segmentation and image generation. For semantic segmentation, mean intersection-over-union (mIoU) and pixel-wise accuracy (acc) are utilized to compare predicted segmentation masks from synthesized images with corresponding ground-truth, and we utilize BiSeNetV2 \cite{yu2021bisenet} to generate segmentation masks. Moreover, Frechet Inception Distance (FID) \cite{heusel2017gans} and learned perceptual image patch similarity (LPIPS) \cite{zhang2018unreasonable} are adopted to evaluate the performance of image generation. In addition, we adopt style relevance (SR) \cite{zhang2020cross}, which uses low-level features of the VGG network \cite{simonyan2014very} to measure the color distance between the synthesized images and ground-truth.

\begin{table}[t]
\centering
\caption{Image manipulation results with random styles on CelebAMask-HQ.}
\begin{tabular}{l||c|c}
\hline
Method        & FID            & LPIPS          \\ \hline \hline
SEAN          & 61.55          & 0.498          \\ \hline
SuperStyleNet & 39.51          & 0.542          \\ \hline
\textbf{Ours}          & \textbf{31.25} & \textbf{0.488} \\ \hline
\end{tabular}
\label{manipulation}
\end{table}

\begin{figure*}[t]
    \centering
    \includegraphics[width=1.0\textwidth]{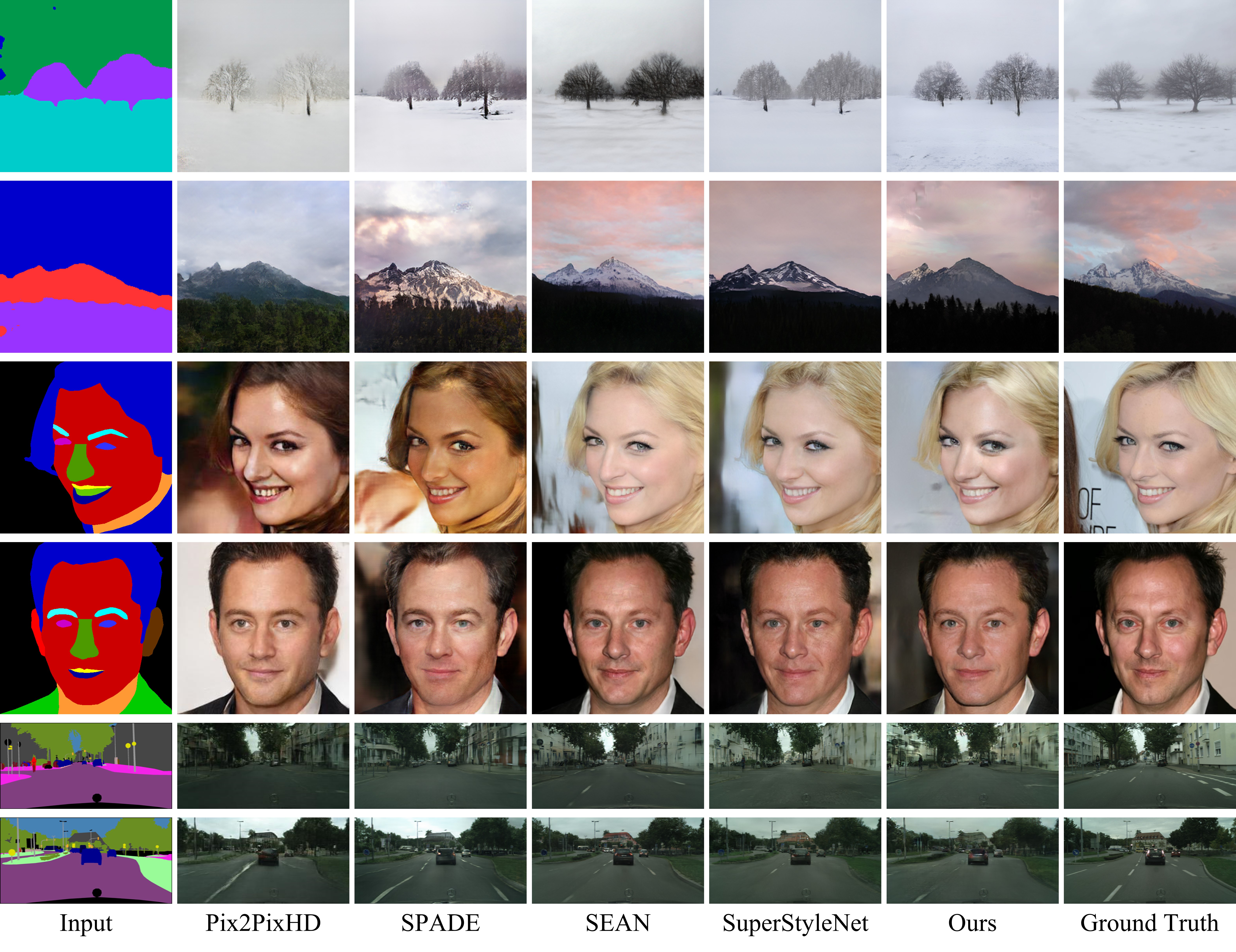}
    \caption{Qualitative comparison with state-of-the-art methods. 1st-2nd row: Landscapes High-Quality. 3-4th row: CelebAMask-HQ. 5-6th row: Cityscapes.}
    \label{qual_comp}
\end{figure*}

\begin{table}[t]
\centering
\scriptsize
\caption{Computational complexity comparison with state-of-the-art methods. \#parameter and time denote the number of parameters and inference time.}
\begin{tabular}{l||c||cc||cc}
\hline
\multicolumn{1}{c||}{\multirow{2}{*}{Method}} & \multicolumn{1}{c||}{\multirow{2}{*}{Backbone}} & \multicolumn{2}{c||}{Landscapes HQ}         & \multicolumn{2}{c}{CelebAMask-HQ}                   \\ \cline{3-6} 
\multicolumn{1}{c||}{}                        & \multicolumn{1}{c||}{} & \multicolumn{1}{c|}{\#parameter}        & time           & \multicolumn{1}{c|}{\#parameter}        & time           \\ \hline \hline
Pix2PixHD                                     & U-Net & \multicolumn{1}{c|}{183M}         & \textbf{0.015} & \multicolumn{1}{c|}{182M}         & \textbf{0.011}   \\ \hline
SPADE                                         & - & \multicolumn{1}{c|}{97M}          & 0.028          & \multicolumn{1}{c|}{92M}          & 0.025          \\ \hline
SEAN                                          & SPADE & \multicolumn{1}{c|}{917M}         & 0.405          & \multicolumn{1}{c|}{266M}         & 0.170          \\ \hline
SuperStyleNet                                 & SPADE & \multicolumn{1}{c|}{912M}         & 3.914          & \multicolumn{1}{c|}{265M}         & 7.073          \\ \hline
\textbf{Ours}                                 & SPADE & \multicolumn{1}{c|}{\textbf{95M}} & 0.106          & \multicolumn{1}{c|}{\textbf{89M}} & 0.037 \\ \hline
\end{tabular}
\label{parameter}
\end{table}

\subsection{Comparisons with state-of-the-art-methods}
\subsubsection{Quantitative and qualitative comparisons}
Table \ref{quan_comp} shows quantitative comparisons with state-of-the-art methods. We do not evaluate semantic segmentation results on the Landscape High-Quality dataset due to the lack of ground-truth. As can be seen from this table, RUCGAN achieves the best performance in style relevance on both Landscape High-Quality and CelebAMask-HQ. Also, RUCGAN shows comparable performance in other evaluation metrics despite utilizing only a single color to predict the style of a semantic region. However, the performance on Cityscapes is lower than other methods since it contains more complex scenes and multiple instances than other datasets. Thus it is difficult to reconstruct details of each style using a singular color as shown in Figure \ref{fail}. Moreover, we compare RUCGAN with state-of-the-art methods in terms of style manipulation on CelebAMask-HQ. To conduct it, we arbitrarily select color of each semantic label to synthesize images. For other methods, style images are randomly selected from a dataset for image synthesis. To minimize the deviation of random selection, we repeat the experiment 5 times and take the average of all the results. As described in Table \ref{manipulation}, RUCGAN achieves the best performance in both FID and LPIPS scores, which indicates that RUCGAN is capable of synthesizing realistic images using random styles.

Furthermore, we also perform qualitative comparisons with state-of-the-art methods. As shown in Figure \ref{qual_comp}, RUCGAN synthesizes high-quality images matched with ground-truth although it merely utilizes a singular color for a style of a semantic region. For the Cityscapes dataset, it shows similar tendencies to the quantitative results.

\begin{table}[t]
\centering
\caption{Ablation study on Landscape High-Quality and CelebAMask-HQ. SCM denotes semantic color mix.}
\begin{tabular}{cc||cc||cc}
\hline
\multicolumn{2}{c||}{Method}                      & \multicolumn{2}{c||}{Landscape HQ}                   & \multicolumn{2}{c}{CelebAMask-HQ}                  \\ \hline
\multicolumn{1}{c|}{P-Norm}       & SCM          & \multicolumn{1}{c|}{FID}           & LPIPS          & \multicolumn{1}{c|}{FID}           & LPIPS          \\ \hline \hline
\multicolumn{1}{c|}{}             &              & \multicolumn{1}{c|}{53.6}          & 0.497          & \multicolumn{1}{c|}{46.3}          & 0.422          \\ \hline
\multicolumn{1}{c|}{$\checkmark$} &              & \multicolumn{1}{c|}{\textbf{20.5}} & 0.399          & \multicolumn{1}{c|}{23.9}          & \textbf{0.280} \\ \hline
\multicolumn{1}{c|}{$\checkmark$} & $\checkmark$ & \multicolumn{1}{c|}{21.3}          & \textbf{0.347} & \multicolumn{1}{c|}{\textbf{21.8}} & 0.282          \\ \hline
\end{tabular}
\label{ablation}
\end{table}

\subsubsection{Computational complexity}
Since each method utilizes different devices, we measure computational complexity employing a single GTX 1080Ti GPU with Pytorch platform for a fair comparison. To evaluate the computational cost, we adopt GMACs (Giga Multiply Accumulate Per Second) that compute the product of two numbers and add it to the accumulator. In this experiment, we perform the generation of a $256\times 256$ synthesized image. Table \ref{parameter} provides comparisons of computational cost and time with state-of-the-art methods on Landscape High-Quality and CelebAMask-HQ. It is clear that RUCGAN achieves the smallest parameters compared with the existing methods. Specifically, the model size of RUCGAN is at least three times less than SEAN \cite{zhu2020sean} and SuperStyleNet \cite{kim2021superstylenet} despite adopting the same backbone.

Overall, RUCGAN is capable of generating realistic synthesized images from segmentation masks with great performance and lightweight architecture.

\begin{figure}[t]
    \centering
    \includegraphics[width=0.45\textwidth]{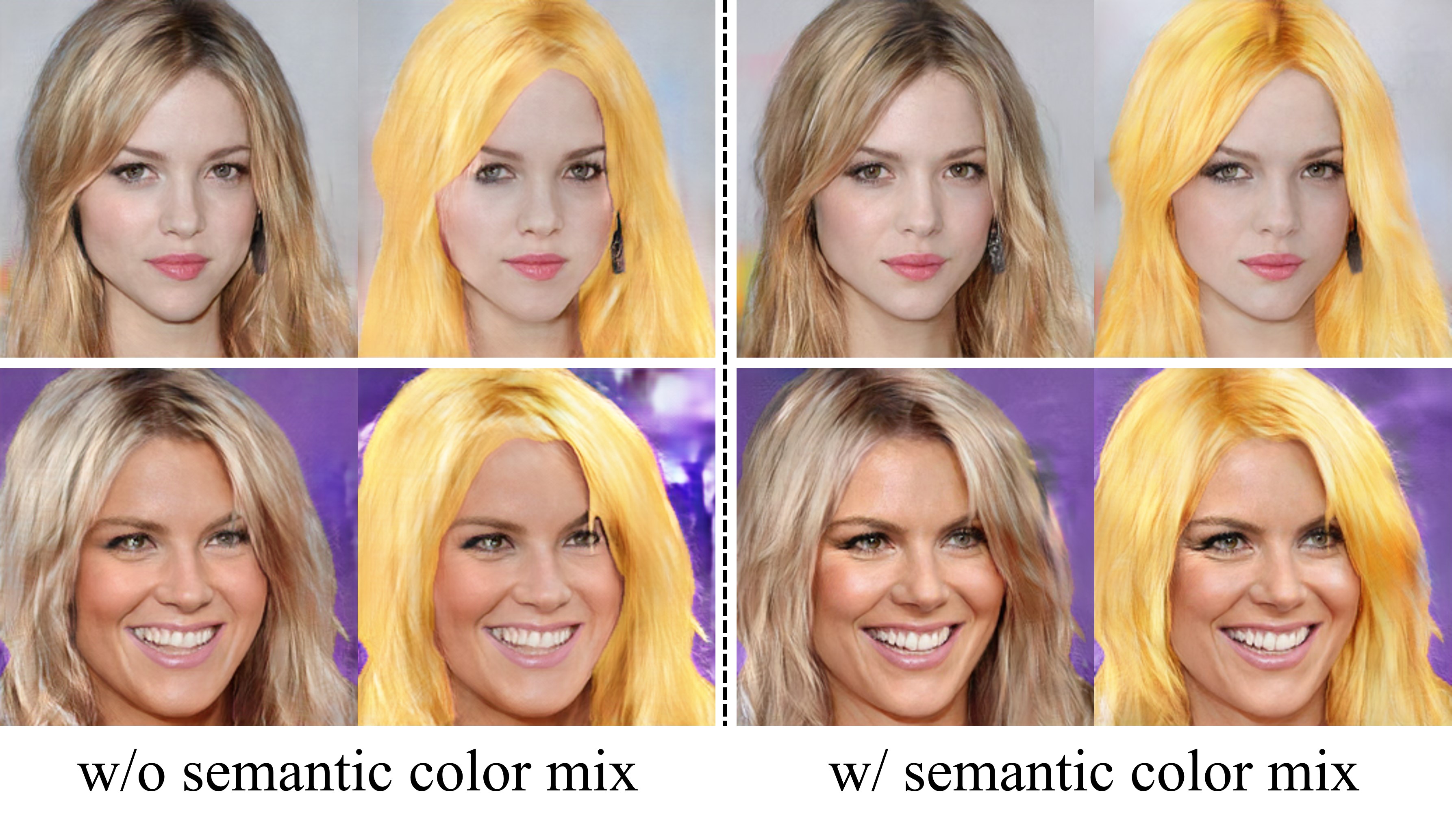}
    \caption{Effects of semantic color mix on image editing. Hair color is changed from left to right images.}
    \label{celeb_scm}
\end{figure}

\subsection{Ablation study}
To demonstrate the effectiveness of our contributions, we perform an ablation study on Landscape High-Quality and CelebAMask-HQ. As shown in Table \ref{ablation}, P-Norm shows a significant improvement on both FID and LPIPS since it uses color information in each semantic label. On the other hand, the addition of semantic color mix does not make much difference to the quantitative results since it does not provide additional information for image synthesis. However, semantic color mix helps RUCGAN to change the style of a specific label independently without affecting other semantic labels. The reason is that semantic color mix conducts semantically-independent color distortion on given semantic regions in the training phase.

\begin{figure}[t]
    \centering
    \includegraphics[width=0.45\textwidth]{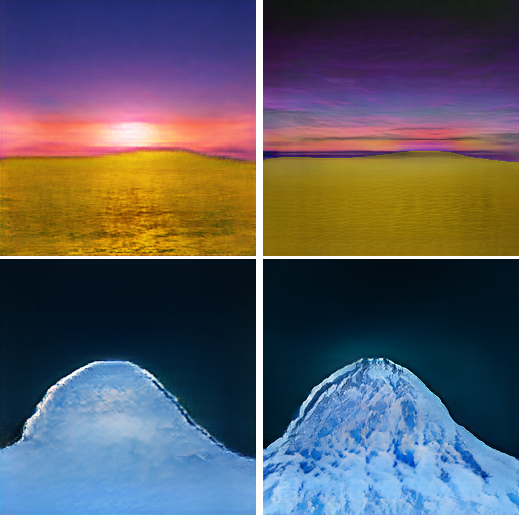}
    \caption{Visualization of RUCGAN results with unusual colors. \textit{(left)} w/o color distortion. \textit{(right)} w/ color distortion. The left images contain a purple sky and a yellow river, and right images show a navy sky and a blue mountain.}
    \label{land_scm}
\end{figure}

In addition, with the help of the color jitter in the semantic color mix, RUCGAN generate realistic synthesized images with unusual colors. As shown in Table \ref{ablation}, semantic color mix dose not show improvement on evaluation metrics; however, it affects local style editing in visual quality. From the examples in Figure \ref{celeb_scm}, we can observe that semantic color mix facilitates RUCGAN to change the style independently without affecting other semantic labels, e.g., a change in hair color does not have a big effect on other semantic regions. The reason is that RUCGAN learns styles of each semantic region from color-distorted and semantically-mixed images. Furthermore,  the color distortion allows RUCGAN to generate realistic synthesized images with unusual colors as shown in Figure \ref{land_scm} which is painted using our UI. As shown in the left images of Figure \ref{land_scm}, both images shows blurring effects and artifacts, and styles of each semantic region are not well-reconstructed. However, these problems are alleviated in the right images with the help of semantic color mix, which proves that RUCGAN is capable of synthesizing photo-realistic images with unusual colors. 

\begin{figure*}[t]
    \centering
    \includegraphics[width=1.0\textwidth]{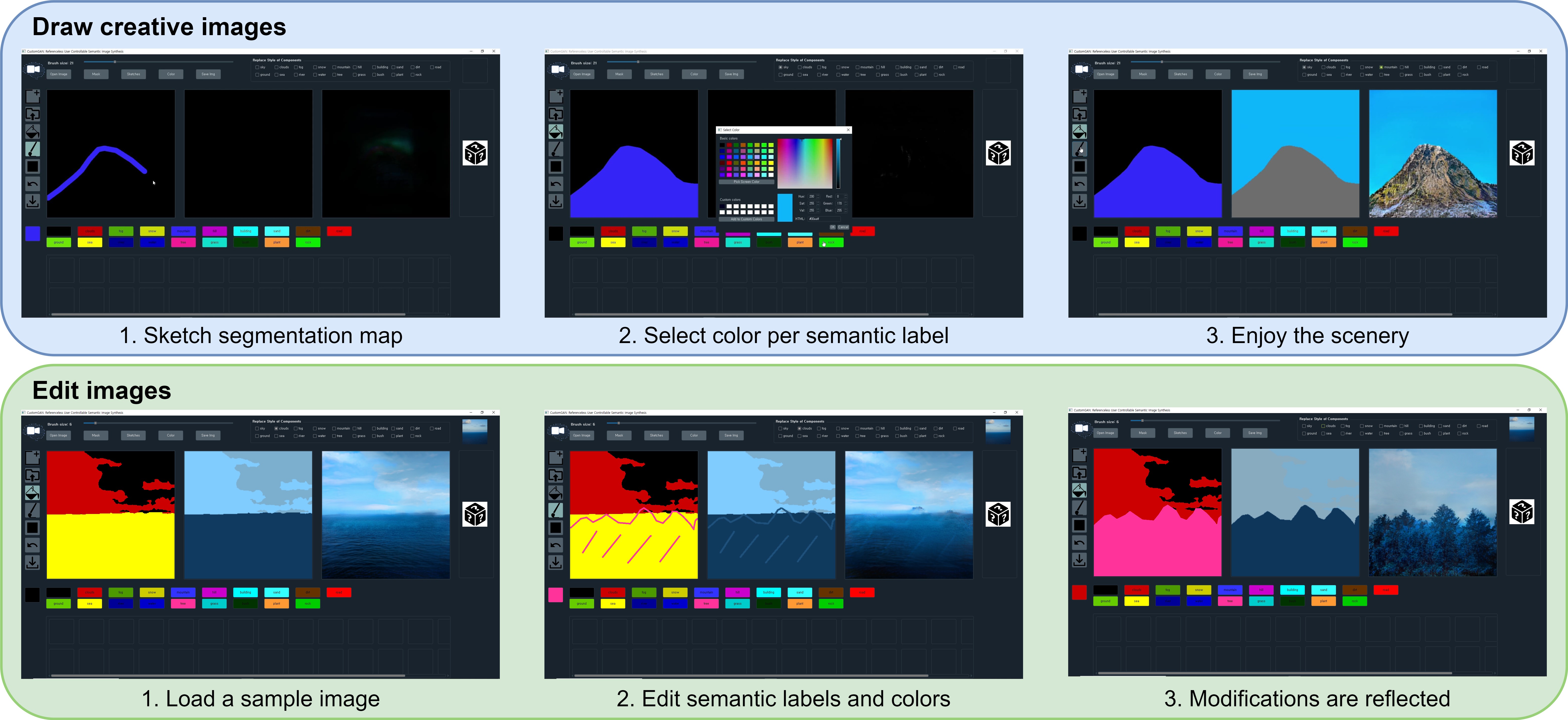}
    \caption{Examples of user interface. \textit{(top)} Draw user-desired images using UI. \textit{(bottom)} Edit local regions of a synthesized image. Screens from left to right in the UI, each showing a segmentation mask, a color-reflected map, and a synthesized image, respectively.}
    \label{ui_img}
\end{figure*}

\begin{figure*}[t]
    \centering
    \includegraphics[width=1.0\textwidth]{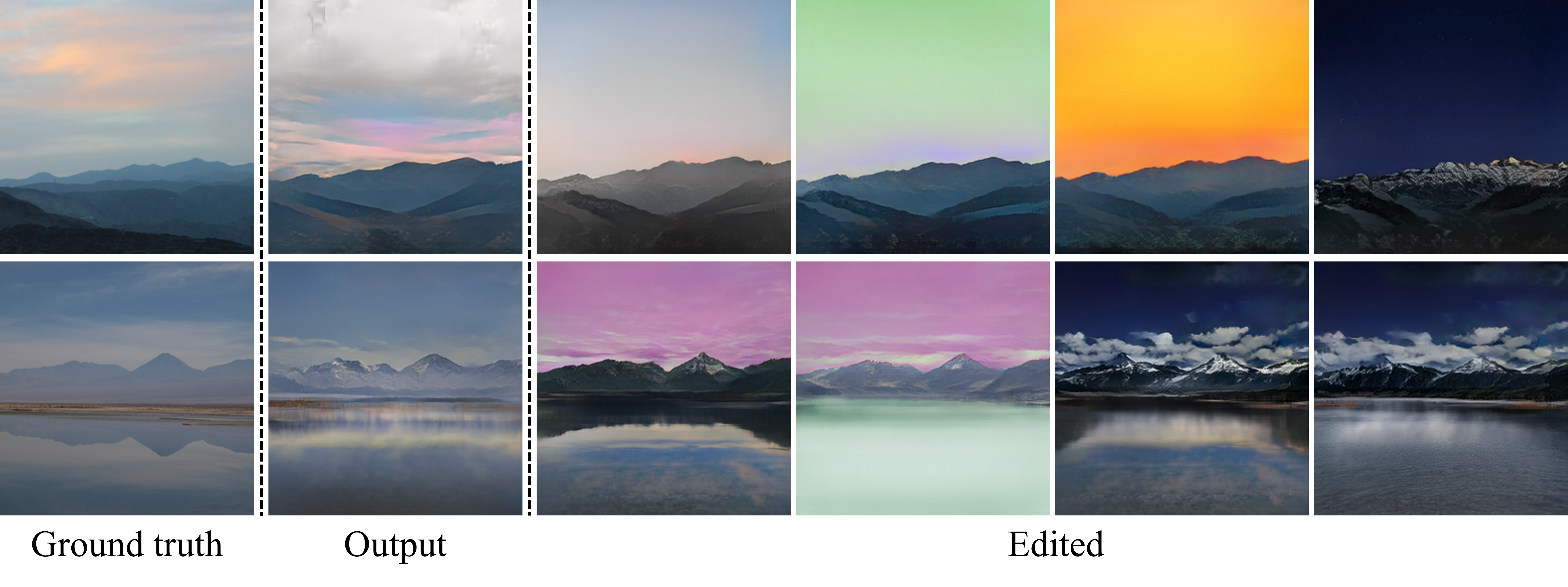}
    \caption{Style editing per semantic region using UI.}
    \label{ui_result}
\end{figure*}

\section{User interface for image synthesis}
The proposed method allows a singular color to represent style information of a specific semantic region. To better display its advantages, we build an interactive user interface. Examples are presented in Figure \ref{ui_img}, which include both image drawing and editing. (1) \textbf{Image drawing}: A segmentation map can be drawn by users using ``brush'' and ``paint'' tools. With the help of P-Norm, styles of each semantic label are mapped from user-selected colors. Then, RUCGAN generates user-customized images based on the segmentation map and the selected colors as shown in the top of Figure \ref{ui_img}. (2) \textbf{Image editing}: RUCGAN is able to edit synthesized images in each semantic region and add additional semantic layout. Firstly, a sample image is loaded from a memory storage. Then, a pretrained segmentation network is used to yield a semantic segmentation map. After that, RUCGAN generates a synthesized image with initial colors from the sample image.  To achieve style editing, users can change the specific semantic label to other labels or select a desired color for a specific semantic label. Moreover, we draw additional segmentation layout in the original one and select a color for its style as shown in the bottom of Figure \ref{ui_img}.

Following the aforementioned steps, users can simply achieve style control for each semantic region without reference images. We provide examples of edited images using UI as shown in Figure \ref{ui_result}.

\section{Conclusion}
We propose a novel network for referenceless user controllable semantic image synthesis, called RUCGAN, to synthesize user-desired images without reference images. Specifically, we introduce palette normalization to learn two modulation parameters from a color palette to inject selected colors into each semantic region as style information. Therefore, a singular color can be utilized as a style. Furthermore, a module termed semantic color mix is integrated in the network, which is able to handle images with unusual styles. To demonstrate the effectiveness of the proposed method, we conduct extensive experiments on various datasets. According to the experimental results, RUCGAN shows comparable performance with the minimum parameters, but it is still challenging to deal with complex scene synthesis. Furthermore, we provide an interactive UI to provide style controllability of the proposed method.

\bibliography{egbib}

\end{document}